\documentclass[conference]{IEEEtran}

\def\BibTeX{{\rm B\kern-.05em{\sc i\kern-.025em b}\kern-.08em
    T\kern-.1667em\lower.7ex\hbox{E}\kern-.125emX}}

\usepackage{amsmath}
\usepackage{amssymb}
\usepackage{amsfonts}
\usepackage{algorithmic}
\usepackage{pzccal}
\usepackage{xcolor}

\newcommand{\Real}{\mathbb{R}}
\newcommand{\Natural}{\mathbb{N}}

\newtheorem{defn}{Definition}
\newtheorem{thm}{Theorem}
\newtheorem{rem}{Remark}

\newtheorem{assume}{Assumption}
\newtheorem{problem}{Problem}

\newcommand{\bfx}{\boldsymbol{x}}

\newcommand{\bfw}{\boldsymbol{w}}

\newcommand{\bfy}{\boldsymbol{y}}

\definecolor{mycolor1}{rgb}{1,0,0}

\definecolor{mycolor2}{rgb}{0.0,0.0,1}

\begin{document}

\title{On Adversarial Examples and Stealth Attacks
in Artificial Intelligence Systems
\thanks{Desmond J. Higham was supported by EP/M00158X/1 from the EPSRC/RCUK Digital Economy Programme,
 EPSRC Programme Grant EP/P020720/1;  Alexander N. Gorban and Ivan Y. Tyukin were supported the Ministry of Science and Higher Education of Russian Federation (Project No. 14.Y26.31.0022).}
}

\author{\IEEEauthorblockN{1\textsuperscript{st} Ivan Y. Tyukin}
\IEEEauthorblockA{\textit{School of Mathematics}\\ \textit{and Actuarial Science} \\
\textit{University of Leicester}\\
and \textit{Norwegian University}\\
\textit{of Science and Technology}\\
and \textit{Saint-Petersburg State}\\
\textit{Electrotechnical University}}\\
{Leicester, LE1 7RH, UK}\\
{and Trondheim, Norway}\\
{and Saint-Petersburg, Russia}\\
{I.Tyukin@le.ac.uk}
\and
\IEEEauthorblockN{2\textsuperscript{nd} Desmond J. Higham}
\IEEEauthorblockA{\textit{School of Mathematics} \\
\textit{University of Edinburgh}}\\
{Edinburgh, EH9 3FD, UK} \\
{d.j.higham@ed.ac.uk}
\and
\IEEEauthorblockN{3\textsuperscript{rd} Alexander N. Gorban}
\IEEEauthorblockA{\textit{School of Mathematics}\\ \textit{and Actuarial Science}\\
\textit{University of Leicester}  \\
and \textit{Lobachevsky University}}\\
{Leicester, LE1 7RH, UK}\\
{and Nizhni Novgorod, Russia} \\
{a.n.gorban@le.ac.uk}
}

\maketitle

\begin{abstract}
In this work we present a formal theoretical framework for assessing and analyzing two classes of malevolent action
towards generic {Artificial Intelligence} (AI) systems.
{Our results apply to general multi-class classifiers that map 
from an 
input space into a decision space,
including artificial neural networks used in deep learning applications.} Two classes of attacks are considered. The first class involves adversarial examples and concerns the introduction of small perturbations of the input data that cause misclassification. The second class, introduced here for the first time and named
{\it stealth attacks}, involves small perturbations to the AI system
itself. Here the perturbed system produces whatever output is
desired by the attacker on a specific small data set, perhaps
even a single input, but performs as normal on a validation
set (which is unknown to the attacker).  

We show that in both cases, i.e., in the case of an attack based on adversarial examples and in the case of a stealth attack, the dimensionality of the AI's decision-making space is a major contributor to the AI's susceptibility. For attacks based on adversarial examples, a second crucial parameter is the absence of local concentrations in the data probability distribution, a property known as Smeared Absolute Continuity. According to our findings, robustness to adversarial examples requires either (a) the data distributions in the AI's feature space to have concentrated probability density functions or (b) the dimensionality of the AI's decision variables to be sufficiently small. 
We also 
show how to construct
stealth attacks on 
high-dimensional AI systems 
that are hard to spot unless the validation set is made exponentially large. \end{abstract}


\maketitle

\section*{Notation}

\begin{itemize}
	\item {$\Real$ denotes the field of real numbers, $\Real_{\geq 0}=\{x\in\Real| \ x\geq 0\}$, and} $\Real^n$ stands for the $n$-dimensional linear real vector space;
	\item $\Natural$ denotes the set of natural numbers;
	\item symbols $\boldsymbol{x} =(x_{1},\dots,x_{n})$ will denote elements of $\Real^n$;
	\item $(\boldsymbol{x},\boldsymbol{y})=\sum_{k} x_{k} y_{k}$ is the inner product of $\boldsymbol{x}$ and $\boldsymbol{y}$, and $\|\boldsymbol{x}\|=\sqrt{(\boldsymbol{x},\boldsymbol{x})}$ is the standard Euclidean norm  in $\Real^n$;
	\item  $\mathbb{B}_n$ denotes the unit ball in $\Real^n$ centered at the origin:
	\[\mathbb{B}_n=\{\boldsymbol{x}\in\Real^n| \ {\|\boldsymbol{x}\|\leq 1}\};\]
	\item  $\mathbb{B}_n(r,\bfy)$  stands for the ball in $\Real^n$ of radius ${r> 0}$ centered at $\bfy$: 
	\[\mathbb{B}_n(r,\bfy)=\{\boldsymbol{x}\in\Real^n| \ {\|\boldsymbol{x}-\bfy\|\leq r}\};\]
	\item  $\mathbb{S}_{n-1}(r,\bfy)$ stands for the $n-1$ sphere in $\Real^n$ that is centered at $\bfy$ and has a radius $r$: {
	\[\mathbb{S}_{n-1}(r,\bfy)=\{\bfx\in\Real^n \ |  \ \|\bfx-\bfy\|=r\};\]}
	\item $V_n$ is the $n$-dimensional Lebesgue measure, and $V_n(\mathbb{B}_n)$ is the volume of unit {$n$}-ball;
\end{itemize}

\section{Background and Motivation}

{The application of 
Artificial Intelligence (AI) and Machine Learning methods 
has produced numerous success stories in recent years
\cite{CBG15,RG16,Schm15}.
Examples where it has been reported that human levels of performance can be  matched or exceeded include
identification of breast cancer \cite{mckinney2020international}, detection of objects hidden from view \cite{caramazza2018neural}, 
mastery of board games \cite{silver16}, 
optimization of new imaging techniques \cite{CMSPE17},
and development of systems for autonomous self-driving cars \cite{bojarski2016end}.} 

{Existing breakthroughs are clearly stimulating further research and encouraging
the broad deployment of such systems in practice.
However, in a field of research where, for example, traffic 
 ``Stop'' signs on the roadside can be misinterpreted  as speed limit signs when  minimal graffiti is added
\cite{physical},
many commentators are asking 
whether current solutions are sufficiently robust, resilient, and trustworthy; and
how such issues should be quantified and addressed. 
Marcus \cite{M18} outlines ten concerns about the current state of deep learning, one of which is that 
``Deep learning thus far works well as an approximation, but its answers often cannot be fully trusted.''
} 

Examples of undesirable, unintended, and unexpected behavior of otherwise sophisticated deep learning systems  raising further questions around the issues of resilience and trustworthiness of data-driven AI systems have been extensively reported and discussed in the literature on {\it adversarial images} \cite{szegedy2013intriguing,harness}.

{
Adversarial images arise when specially chosen perturbations, effectively imperceptible to the human eye, cause misclassification in an AI system, or, indeed, 
simultaneously across a range of AI systems.
}
The existence of 
adversarial images illustrates the risks associated with the deployment of data-driven neural network-based decision-making 
{and raises important questions around
responsible research and innovation (RRI) \cite{Gr16,Dav17}.
There are now many constructive 
approaches 
for the generation of adversarial attacks; for example, 
\cite{kurakin2016adversarial,deepfool,sparsefool,squareattack,deepfake2,bb,SVK17}.
On the other hand, techniques that aim to identify or guard against such attacks
have also been developed; for example, 
\cite{Pars17,CAH18,athalye2018obfuscated,Croce2020Provable,robust, yuan2019adversarial}, leading to 
a version of conflict escalation where attack and defence strategies become
increasingly ingenious. 
}

{
Against this backdrop, the work in 
\cite{shafahi2018adv} 
looks at a higher-level question: are there 
fundamental reasons that make adversarial examples difficult to thwart?} 
The authors developed arguments based on various versions of the isoperimetric inequality to determine a set of conditions under which adversarial examples occur with probability close to one in a very general setting (see \cite{shafahi2018adv} for further details).

{
In this work, we use a different set of tools to derive  alternative conditions under which  
the existence of adversarial examples is essentially unavoidable
for general classifiers.}
In addition, 
we
introduce a
second type of risk, relating to malicious, targeted behavior
that we refer to as a \emph{stealth attack}.
In this scenario, an attacker (who may, for example, be
a mischievous,  disgruntled, malevolent or corrupt member of a large software development team)
has access to the actual code implementing the AI system.
Such an attacker is capable
of changing, adding or replacing a single or a small number of nodes with the aim of altering the behavior of the system. To evade detection,
the perturbed system must show little if any deviation from the nominal system's expected performance on some finite verification set $\mathcal{V}$, making the attack
 transparent to the AI's owners and users. At the same time, on a data set or even a single input  
$\bfx'$ that is known only to the attacker, the system must generate a response which the attacker desires but which is different from the nominal system's output. (So, for example, there may be a particular image whose classification the attacker
wishes to override.)

If the verification set $\mathcal{V}$ is available to the attacker and the attacker is allowed to change a significant portion of the nominal AI system (e.g., parameters and connections of neurons in the network) then {it is technically plausible and operationally simple to execute} such an attack by re-training. Large systems in which the total number of parameters of the altered part exceeds  the cardinality of $\mathcal{V}\cup \bfx'$ {are particularly vulnerable} to {alterations} of this type. Indeed, it is well-known that $n+1$ {generic} points in $\Real^n$ are linearly separable. 
Experiments in \cite{zhang2016understanding} showed that simple shallow yet sufficiently large neural networks may achieve perfect finite sample expressivity as soon as the number of parameters exceeds the number of data points  ({cf.} \cite{cover1965}).

We have in mind the more challenging case when i) the set $\mathcal{V}$ and its cardinality is unknown to the attacker and ii) the attacker may {change} only a single element (albeit with its weights and parameters) in the system.

\section{General Framework}

{We will study
both adversarial examples and stealth attacks in a single, generic, setting.}
We suppose that the system is modeled by a map
\begin{equation}\label{eq:classification_map}
\mathcal{F}: \ {\mathbb{B}_n} \rightarrow \Real.
\end{equation}
The map may represent a multi-class classifier, {implemented e.g., by} a neural network, defined on a set $\Phi\subset {\mathbb{B}_n}$ of the feature vectors $\bfx\in\Phi$. The nature and the origin of the feature vectors and the map itself are not important for our analysis. The map can be viewed as a transformation modelled by one or a few fully connected layers inside a deep neural network; it may also describe the entire input-output behavior of the system.  What is important, however, is that the feature vectors $\bfx$ are elements of a high-dimensional vector space $\mathbb{R}^n$.

Using this model, we formally analyze the inevitability of both adversarial examples and stealth attacks.
With respect to the problem of adversarial examples (Theorem \ref{theorem:adversarial_examples}), we formulate a relationship between a given classifier and statistical properties of the data (Assumption \ref{assume:class_density}) {that leads} inevitably to the existence of adversarial examples. A key element of these conditions is the Smeared Absolute Continuity (SmAC) property of the probability distribution introduced in \cite{gorban2018correction}. A similar condition is imposed in \cite{shafahi2018adv} in the form of the assumption of an upper bound for the probability density function. Here, however, we do not require that the latter property holds for the entire distribution. If $n$ is sufficiently large then for the existence of an $(\varepsilon+\Delta)$-adversarial example ($\varepsilon$ may be chosen arbitrarily small) it is sufficient that 
\begin{description}
\item[i)] the SmAC condition holds in  some ball of non-zero measure, and
\item[ii)] for any point on the boundary of that ball there is an element of a different class within distance $\Delta$. 
\end{description}
We also provide an explicit estimate of the dimension $n$ at which such examples become probable.

{The new concept of a stealth attack, where an opponent  modifies a small part of the backbone system in a way that impacts only specific inputs, is formalized 
(\ref{eq:adversarial_constraints}).}
Our results show that 
stealth attacks 
 are surprisingly easy to construct for large enough  $n$. 
 In particular, we find that if the cardinality $M$ of the verification set $\mathcal{V}$ is smaller than $2^n$ then these attacks can be produced by a modification of a single node in the system and without any knowledge of the verification data (Theorem \ref{theorem:adversarial_changes}). 

The rest of the manuscript is organized as follows: in Section \ref{sec:adversarial_examples} we quantify probabilities of adversarial examples for a broad class of data distributions satisfying the SmAC condition, Section \ref{sec:adversarial_changes} presents conditions and possible scenarios for  stealth attacks, and Section \ref{sec:conclusion} concludes the paper.

\section{Adversarial examples}\label{sec:adversarial_examples}

Consider a standard multi-class classification problem in which  each element $\bfx\in\Phi$  is  associated with a label $\mathpzc{l}\in\mathcal{L}$ from a finite set $\mathcal{L}$ of labels. We assume that the pairs $(\bfx,\mathpzc{l})$ are drawn from some probability distribution with the  corresponding probability density function:
\[
p: \ {\mathbb{B}_n}\times \mathcal{L} \rightarrow \Real_{\geq 0}.
\]
The distribution as well as the probability density functions are supposed to be {\it unknown}  but their existence is assumed. The backbone/legacy AI system is hence a classifier which for a given $\bfx\in\Phi$ aims at predicting its label $\mathpzc{l}$.

\begin{defn}\label{definition:adversarial_example} For the given classification map $\mathcal{F}$,  an element $\bfx\in{\mathbb{B}_n}$ admits a $\delta$-adversarial example $\bfy(\bfx)$ if
\[
\mathcal{F}(\bfx)\neq \mathcal{F}(\bfy(\bfx)) \ \mbox{and} \ \|\bfx-\bfy(\bfx)\|\leq \delta, \ \bfy(\bfx)\in{\mathbb{B}_n}.
\]
\end{defn}

In what follows we will determine a set of conditions on the classifier map $\mathcal{F}$ and the data distribution for which adversarial examples exist and the probability of their occurrence is non-zero and sometimes could be even exponentially ``close'' to $1$ with respect to  $n$.

Let $A$ be an element of the label set $\mathcal{L}$. 
We denote

\begin{eqnarray}\label{eq:probability_definitions}
& & p_A(\bfx)=p(\bfx|\mathpzc{l}=A), \ P(A)=\int_{{\mathbb{B}_n}} p(\bfx,A) d\bfx, \nonumber \\
& & p(\bfx|\mathpzc{l}=A)=\frac{p(\bfx,A)}{P(A)}.
\end{eqnarray}

\begin{assume}\label{assume:class_density} There exists a label $A\in\mathcal{L}$ and an associated set $\mathcal{C}_A\subset{\mathbb{B}_n}$, a  {number $r_A\in(0,1)$},  a vector $\bfx_A\in{\mathbb{B}_n}$, {a positive constant $C>0$, and a  number $\nu\in(0,1]$} such that

\begin{itemize}
\item[A1) ] The set $\mathcal{C}_A$ is contained in $\mathbb{B}_n(r_A,\bfx_A)$.
\item[A2) ] $\mathcal{F}(\bfx)=A$ for all $\bfx\in\mathcal{C}_A$, and there is a $\Delta>0$ such that  for any  $\bfx\in \mathbb{S}_{n-1}(r_A, \bfx_A)$ there exists a $\bfy(\bfx)$:
\[
\mathcal{F}(\bfy(\bfx))\neq A, \quad \|\bfy(\bfx)-\bfx\|\leq \Delta.
\]
\item[A3) ] The probability density function $p_A$   satisfies
\begin{eqnarray}
&& p_A(\bfx)\leq \frac{C}{{V_n(\mathbb{B}_n})} \frac{1}{r_A^n} \ \mbox{for all} \ \bfx\in\mathbb{B}_n(r_A,\bfx_A),\nonumber \\
&& {\mbox{and}} \int_{\mathcal{C}_A}p_A(\bfx)d\bfx \geq \nu >0. \label{eq:class_density}
\end{eqnarray}

\end{itemize}
\end{assume}

Conditions $A1$ -- $A3$ in Assumption \ref{assume:class_density} formalize a relationship between the given classification map $\mathcal{F}$ and statistical properties of the pair $(\bfx,\mathpzc{l})$ which, as we shall see later, lead to the risk of emergence of adversarial examples. In particular,
Assumption \ref{assume:class_density} ensures that
\begin{itemize}
\item The probability that the event $\bfx\in\mathcal{C}_A, \mathpzc{l}=A$ occurs is at least $P(A)\nu$, and the corresponding conditional probability density $p_A$ satisfies a form of the Smeared Absolute Continuity condition in the domain $\mathcal{C}_A$ \cite{gorban2018correction} (condition $A3$).

\item Any $\bfx$ from the set $\mathcal{C}_A$ is  interpreted as an element of class $A$ by the map $\mathcal{F}$, and a $\Delta$-neighborhood of any element $\bfx$ on the boundary of the set $\mathbb{B}_n(r_A,\bfx_A)\supset\mathcal{C}_A$ contains at least one element $\bfy(\bfx)$ to which the map $\mathcal{F}$ assigns a label that is different from $A$ (condition $A2$). The latter part of the condition will obviously hold if
\[
\begin{split}
& \mathcal{F}(\bfx)\neq A  \ \mbox{for all}  \\
& \ \ \ \ \ \ \ \ \ \bfx\in\mathbb{B}_n(r_A+\Delta, \bfx_A)\cap {\mathbb{B}_n}\setminus \mathbb{B}_n(r_A,\bfx_A).
\end{split}
\]
\item A non-empty set for which the above properties hold exists (the set $\mathcal{C}_A)$ and is in the interior of {some $n$-ball in} ${\mathbb{B}_n}$ (condition $A1$).
\end{itemize}

Under these conditions the following statement holds.

\begin{thm}\label{theorem:adversarial_examples} Consider a classification map $\mathcal{F}$ and a probability distribution with probability density function $p$  satisfying Assumption \ref{assume:class_density}. Let {a} sample $(\bfx,\mathpzc{l})$ be drawn from this distribution and let $\varepsilon$ be chosen arbitrarily in  $(0,r_A)$. Then the probability that $\bfx$ admits an $(\varepsilon+\Delta)$-adversarial example is at least
\begin{equation}\label{eq:adversarial_example_probability}
P(A)\max\left\{\nu - C \left(1-\frac{\varepsilon}{r_A}\right)^n,0\right\}.
\end{equation}
\end{thm}
{\it Proof of Theorem \ref{theorem:adversarial_examples}} Let us fix an $0<\varepsilon < r_A$ and let $P^\ast$ be the probability of the event
\[
\bfx\in\mathbb{B}_n(\bfx_A, r_A)\setminus {B}_n(\bfx_A, r_A-\varepsilon), \ \mathpzc{l}=A.
\]
Then according to Assumption \ref{assume:class_density} (condition $A2$) and Definition \ref{definition:adversarial_example}, the probability that a $\Delta+\varepsilon$ adversarial example exists for the given classifier is $P^\ast$. The probability $P^\ast$ can be estimated as
\[
P^\ast= P(A) P(\bfx\in\mathbb{B}_n(\bfx_A, r_A)\setminus {B}_n(\bfx_A, r_A-\varepsilon)| \mathpzc{l}=A) .
\]
Consider
\begin{eqnarray}
&&P(\bfx\in{B}_n(\bfx_A, r_A-\varepsilon)|\mathpzc{l}=A) \nonumber \\
&&=\int_{ {B}_n(\bfx_A, r_A-\varepsilon)}p(\bfx|\mathpzc{l}=A)d\bfx \nonumber \\
&&=\int_{ {B}_n(\bfx_A, r_A-\varepsilon)}p_A(\bfx)d\bfx. \nonumber
\end{eqnarray}
According to (\ref{eq:probability_definitions}) and (\ref{eq:class_density}),
\begin{eqnarray}
& &\int_{ {B}_n(\bfx_A, r_A-\varepsilon)}p_A(\bfx)d\bfx \leq  \int_{ {B}_n(\bfx_A, r_A-\varepsilon)} \frac{C}{{V_n(\mathbb{B}_n} )r_A^n}d\bfx\nonumber\\
& & C \frac{(r_A-\varepsilon)^n}{r_A^n} = C\left(1-\frac{\varepsilon}{r_A}\right)^n. \nonumber
\end{eqnarray}
Using conditions $A1$ and $A3$ from Assumption \ref{assume:class_density}, we can obtain the following estimate
\[
\begin{split}
& P(\bfx\in\mathbb{B}_n(\bfx_A, r_A)\setminus {B}_n(\bfx_A, r_A-\varepsilon)| \mathpzc{l}=A)=\\
& P(\bfx\in\mathbb{B}_n(\bfx_A, r_A)|\mathpzc{l}=A)-\\
& P({B}_n(\bfx_A, r_A-\varepsilon)| \mathpzc{l}=A)\geq \nu - C \left(1-\frac{\varepsilon}{r_A}\right)^n.
\end{split}
\]
The value of $P^\ast$ can now be estimated from below as
\[
P(A)\left( \nu - C \left(1-\frac{\varepsilon}{r_A}\right)^n\right),
\]
and hence the statement follows $\square$.

Using the well-known inequality 
\[
(1-x)^{1/x}<e^{-1}, \quad x\in(0,1),
\]
 the following exponential lower bound estimate for (\ref{eq:adversarial_example_probability}) holds:
\[
P(A)\max\left\{ \nu - C \exp \left(-\frac{n\varepsilon}{r_A}\right),0\right\}.
\]

\begin{rem} \normalfont According to Theorem \ref{theorem:adversarial_examples}, if the classifier and the data probability distribution satisfy Assumption \ref{assume:class_density}, then  $(\varepsilon+\Delta$)-adversarial examples are expected to occur if the dimensionality $n$ of the feature space is  sufficiently large:
\begin{equation}\label{eq:critical_dimension}
n > ({\log \nu - \log C})\left[{\log \left(1-\frac{\varepsilon}{r_A}\right)}\right]^{-1}.
\end{equation}
Moreover, if $C$ is independent of $n$, then the probability that the data sample admits a $(\varepsilon+\Delta$)-adversarial example  approaches $P(A)\nu$ exponentially fast with dimension $n$.
\end{rem}

\begin{rem}\normalfont For classifiers operating in dimensions satisfying (\ref{eq:critical_dimension}) one can now easily derive a bound on the probability of occurrence {of} an $(\varepsilon+\Delta)$- adversarial example {in a sample of $N$   i.i.d. random data points}. In particular,  under the assumptions of Theorem \ref{theorem:adversarial_examples}, the probability that at least one  $(\varepsilon+\Delta)$-adversarial example occurs {this sample} is not smaller than
\[
1-\left[1-P(A)\left( \nu - C \left(1-\frac{\varepsilon}{r_A}\right)^n\right)\right]^N.
\]
\end{rem}

\section{Stealth Attacks to the backbone AI}\label{sec:adversarial_changes}

The susceptibility of decision-making in AI systems operating in high-dimensional space to small adversarial perturbations of the data is just one facet of the larger topic of robust, resilient, and ultimately verifiable AI performance.
In this subsection we formally define and study the related but distinct issue of stealth attacks.

To set-up our framework, consider the classification map (\ref{eq:classification_map})
\[
\mathcal{F}:{\mathbb{B}_n}\rightarrow \Real
\]
modelling the backbone AI system. In addition to this map, consider
\begin{equation}\label{eq:adversarial_map}
\begin{split}
& \mathcal{F}_{a}:{\mathbb{B}_n}\times \Theta \rightarrow \Real \\
& \mathcal{F}_{a}(\cdot,\boldsymbol{\theta})=\mathcal{F}(\cdot) + \mathfrak{A}(\cdot,\boldsymbol{\theta}),
\end{split}
\end{equation}
where the term
\[
\mathfrak{A}: {\mathbb{B}_n}\times \Theta \rightarrow \Real
\]
{models} a stealth attack on the original backbone system $\mathcal{F}$, and $\Theta\subset \Real^m$ is an associated set of parameters.

A case of significant practical interest arises when the term $\mathfrak{A}$ can be expressed using {just a single} Rectified Linear Unit (ReLU function),   \cite{hahnloser2000digital} (see, for example, \cite{gorban1996neural}
or
\cite{HHdl2019}
for information
regarding basic nonlinear elements)
\begin{equation}\label{eq:ReLU}
\begin{split}
& \mathfrak{A}(\cdot,(\bfw,b))= D \mathrm{ReLU}((\cdot,\bfw)-b), \\
& \mathrm{ReLU}(s)=\max \{s,0\}
\end{split}
\end{equation}
or a sigmoid
\begin{equation}\label{eq:sigmoid}
\begin{split}
& \mathfrak{A}(\cdot,(\bfw,b))= D \sigma((\cdot,\bfw)-b), \\
&  \sigma(s)=\frac{1}{1+\exp(- s)},
\end{split}
\end{equation}
with $D>0$ being a positive constant. It is convenient to denote
\[
\mathfrak{A}(\cdot,(\bfw,b))=D g ((\cdot,\bfw)- b),
\]
where the function $g$ is either $\mathrm{ReLU}$ or sigmoid, depending on the case, {and $(\bfw,b)=\boldsymbol{\theta}$ are its relevant parameters}. We are now ready to formally introduce the following stealth attack problem

\begin{problem}[Stealth Attack on $\mathcal{F}$]\label{problem:adversarial_change} \normalfont Consider a classification map $\mathcal{F}$ defined by (\ref{eq:classification_map}) and modelling a backbone AI. Suppose that an owner of the AI system or a network has a finite  validation or verification set
\[
\mathcal{V}\subset{\mathbb{B}_n}.
\]
The validation set $\mathcal{V}$ is kept secret and is assumed to be {\it unknown} to an attacker. The cardinality of $\mathcal{V}$ is bounded from above by some constant $M$, and this bound is known to the attacker.

The attacker seeks to modify the map $\mathcal{F}$ and replace it by $\mathcal{F}_a$ constructed in accordance with (\ref{eq:adversarial_map}), (\ref{eq:ReLU}) or (\ref{eq:adversarial_map}), (\ref{eq:sigmoid}) and such that for some given $\varepsilon>0$, $\Delta>0$ and an element $\bfx'\in{\mathbb{B}_n}$, known to the attacker but unknown to the owner of the map $\mathcal{F}$, the following properties hold:
\begin{equation}\label{eq:adversarial_constraints}
\begin{split}
&\|\mathcal{F}(\bfx)-\mathcal{F}_a(\bfx,(\bfw,b))\|\leq \varepsilon \ \forall \ \bfx\in\mathcal{V}\\
& \mathcal{F}_a(\bfx',(\bfw,b))=\mathcal{F}(\bfx') + \Delta.
\end{split}
\end{equation}
\end{problem}
In words, the stealth attack has an imperceptible effect on the validation set, {since $\varepsilon>0$ can be made arbitrarily small}, but makes the desired modification {of the backbone AI (with arbitrarily large $\Delta>0$)} for the target input $\bfx'$.

We say that $\mathfrak{A}$ is a solution of this problem if it satisfies (\ref{eq:adversarial_constraints}). The next statement provides an efficient mechanism for constructing such solutions.

\begin{thm}\label{theorem:adversarial_changes} Consider Problem \ref{problem:adversarial_change}, and let $\bfx'$ be a vector that is randomly drawn from the equidistribution in ${\mathbb{B}_n}$. Then the probability that
\begin{equation}\label{eq:adversarial_solution}
\begin{split}
&\mathfrak{A}(\cdot, (\kappa\bfx', b))=D g((\cdot,\kappa \bfx') - b), \\
&b=\kappa \left(\frac{1+\gamma}{2}\right) \|\bfx'\|^2,
\end{split}
\end{equation}
where $\kappa$ and $D$ are chosen so that
\begin{equation}\label{eq:adversarial_solution_parameters}
\begin{split}
&D g \left(-\kappa \frac{1-\gamma}{2} \|\bfx'\|^2 \right)  \leq \varepsilon \ \mbox{and} \\
&D g \left(\kappa \frac{1-\gamma}{2} \|\bfx'\|^2 \right)  \geq \Delta, \ \gamma\in(0,1),
\end{split}
\end{equation}
is a solution of Problem \ref{problem:adversarial_change} is at least
\[
1-M\left(\frac{1}{2\gamma}\right)^n.
\]
\end{thm}

{\it Proof of Theorem \ref{theorem:adversarial_changes}}. Let us pick $\gamma\in(0,1)$ and let $\bfx'$  be such that
\[
\gamma (\bfx',\bfx')=\gamma\|\bfx'\|^2 > (\bfx',\bfx_i), \ \mbox{for all} \ \bfx_i\in\mathcal{V}.
\]
Set
\begin{eqnarray*}
\bfw &=& \kappa \bfx',  \ \kappa>0, \\
b & =& \kappa \left(\frac{1+\gamma}{2}\right) \|\bfx'\|^2,
\end{eqnarray*}
and observe that
\[
\mathfrak{A}(\cdot,(\bfw,b))=D g\left(\kappa\left( (\cdot,\bfx') - \left(\frac{1+\gamma}{2}\right) \|\bfx'\|^2 \right)\right),
\]
where we recall that $g$ is either ReLU or sigmoid. Consider
\[
\|\mathcal{F}(\bfx_i)-\mathcal{F}_a(\bfx_i,(\bfw,b))\|= |\mathfrak{A}(\cdot,(\bfw,b))|.
\]
Since the function $g$ is monotone,
\[
|\mathfrak{A}(\bfx_i,(\bfw,b))|\leq D g\left(- \kappa \left(\frac{1-\gamma}{2} \|\bfx'\|^2 \right)\right) \ \forall \ \bfx_i\in\mathcal{V}.
\]
Denote
\[
z = \frac{1-\gamma}{2} \|\bfx'\|^2
\]
and pick the values of $D$ and $\kappa$ so that
\[
D g (-\kappa z)  \leq \varepsilon \ \mbox{and} \
D g (\kappa z )  \geq \Delta.
\]
Given that $\mathrm{ReLU}(s)=0$ for all $s\leq 0$ and that the sigmoidal function is strictly increasing with $g(0)\neq 0$, such choice is always possible.

Finally, let $\bfx'$ be drawn from the equidistribution in ${\mathbb{B}_n}$.   Then the probability that
\[
\gamma (\bfx',\bfx') > (\bfx',\bfx_i), \ \mbox{for all} \ \bfx_i\in\mathcal{V}
\]
is at least
\[
1-M\left(\frac{1}{2\gamma}\right)^n.
\]
(see Proposition 1 of \cite{gorban2018correction}). This completes the proof. $\square$

\begin{rem}\normalfont If $g=\mathrm{ReLU}$ then the value of $\varepsilon$ in Theorem \ref{theorem:adversarial_changes} can be set to $0$ which in turn implies that the stealth map $\mathcal{F}_a$ is indistinguishable from $\mathcal{F}$ on the verification set $\mathcal{V}$:
\[
\mathcal{F}_a(\bfx)=\mathcal{F} \ \forall \ \bfx\in\mathcal{V}.
\]
\end{rem}

\begin{rem}\normalfont The statement of Theorem \ref{theorem:adversarial_changes} can be adjusted to include the class of functions $g$:
\[
\lim_{s\rightarrow -\infty } g(s)=0, \ \lim_{s\rightarrow \infty } g(s)=0, \ g(0)=1.
\]
In this case the value of $b$ in (\ref{eq:adversarial_solution}) should change to 
\[
b=\kappa\|\bfx'\|^2
\]
 and condition (\ref{eq:adversarial_solution_parameters}) will need to become
\[
D g \left(-\kappa (1-\gamma) \|\bfx'\|^2 \right)  \leq \varepsilon \ \mbox{and} \
D  \geq \Delta, \ \gamma\in(0,1).
\]
This extends the results to bell-shaped functions $g$ such as the Gaussian and also opens possibilities to use general sigmoidal functions $\sigma$ to construct such $g$:
\[
g(s)=\frac{\sigma(s)-\sigma(s+a)}{\sigma(0)-\sigma(0+a)}, \quad  \sigma(0)-\sigma(0+a)\neq 0.
\]
\end{rem}

\section{Conclusion}\label{sec:conclusion}

In this work we set up a formal framework for analyzing two classes of malevolent action
towards generic AI systems. These systems include neural networks but generally could be of a rather arbitrary type.
The first class, adversarial examples, concerns small perturbations of the input data that
cause misclassification. Such perturbations have been widely studied in recent years,
mostly from an empirical perspective.
The second class, introduced here for the first time and
named
stealth attacks, involve
small perturbations to the AI system itself.
Here the perturbed system produces whatever output is desired by the attacker on a specific small data set, perhaps even a single input, but
performs as normal on
a validation set (which is unknown to the attacker).

In both cases, we identified the dimensionality of the AI's decision-making space
as a major factor in its susceptibility.

With regard to adversarial examples, a second crucial aspect influencing the risk of adversarial attacks is the absence or presence of local concentrations in  the data probability distribution (Smeared Absolute Continuity condition). According to our findings, a robust system should either have concentrated probability density functions or its dimensionality must be reduced to avoid the effects of the measure concentration.

Concerning stealth attacks on the backbone AI, we note that systems with $\mathrm{ReLU}$ activation functions are particularly prone to adversarial modifications which are hard to spot without resorting to exponentially large in dimension, $2^n$, verification sets. Single-node adversarial alterations involving differentiable activation functions may need to have large Lipschitz constants (i.e., the values of $\kappa, D$ in  Theorem \ref{theorem:adversarial_changes}). Lipschitz constants calculated over a data sample have been used extensively as an indicator of network quality  \cite{gorban1996neural} (the smaller the better). Here we have shown that
these are not only mere quality indicators; large Lipschitz constants in networks and systems with differentiable activation functions are also consistent with
susceptibility to stealth attack.

Many relevant questions, however, remain. In particular, we did not consider
here probabilities of noise-induced misclassifications. We also did not try to produce the tightest possible probability estimates. Addressing these, and related issues,
will be the focus of future work.

\section*{Acknowledgement}

Desmond J. Higham was supported by EP/M00158X/1 from the EPSRC/RCUK Digital Economy Programme and 
 EPSRC Programme Grant EP/P020720/1;  Alexander N. Gorban and Ivan Y. Tyukin were supported the Ministry of Science and Higher Education of Russian Federation (Project No. 14.Y26.31.0022).

\bibliographystyle{IEEEtran}
\bibliography{adversarial_concentration}

\end{document}